# Structured Reachability Analysis for Markov Decision Processes

**Craig Boutilier***
Department of Computer Science
University of British Columbia
Vancouver, BC, Canada V6T 1Z4
cebly@cs.ubc.ca

**Ronen I. Brafman**[†]
Department of Math and CS
Ben-Gurion University
Beer Sheva, Israel 84105
brafman@cs.bgu.ac.il

**Christopher Geib**[‡]
Honeywell Technology Center
MN65-2600, 3660 Technology Dr.
Minneapolis, MN, US A 55418
geib@htc.honeywell.com

## Abstract

Recent research in decision theoretic planning has focussed on making the solution of Markov decision processes (MDPs) more feasible. We develop a family of algorithms for *structured reachability analysis* of MDPs that are suitable when an initial state (or set of states) is known. Using compact, structured representations of MDPs (e.g., Bayesian networks), our methods, which vary in the tradeoff between complexity and accuracy, produce structured descriptions of (estimated) reachable states that can be used to eliminate variables or variable values from the problem description, reducing the size of the MDP and making it easier to solve. One contribution of our work is the extension of ideas from GRAPHPLAN to deal with the distributed nature of action representations typically embodied within Bayes nets and the problem of correlated action effects. We also demonstrate that our algorithm can be made more *complete* by using *k*-ary constraints instead of binary constraints. Another contribution is the illustration of how the compact representation of reachability constraints can be exploited by several existing (exact and approximate) abstraction algorithms for MDPs.

## 1 Introduction

While Markov decision processes (MDPs) have proven to be useful as conceptual and computational models for decision theoretic planning (DTP), there has been considerable effort devoted within the AI community to enhancing the computational power of these models. One of the key drawbacks of classic algorithms such as policy iteration [16] or value iteration [1] is the need to explicitly "sweep through" state space, making these techniques impractical for realistic problems. Recent research on the use of MDPs for DTP has focussed on methods for solving MDPs that avoid explicit state enumeration while constructing optimal or approximately optimal policies. These include the use of function approximators for value functions [2], aggregation and abstraction techniques [5, 6, 12, 8], reachability analysis [9], and decomposition techniques [11].

In this paper we address the problem of integrating reachability considerations into the construction of *abstract MDPs*. In particular, we develop techniques whereby knowledge of an initial state (or initial conditions) and the concomitant reachability considerations influence the abstractions produced for an MDP, forming what is termed by Knoblock [17] a *problem specific abstraction*. We assume that the MDP is described in terms of random variables using *dynamic Bayes nets* (DBNs) [6]; we also assume an initial state has been given. Given this representation, several useful (exact and approximate) abstraction techniques can be employed to solve an MDP without explicit state space enumeration [5, 6, 12].[1] These all rely on the identification of conditions under which a variable can influence a value function or action choice, and can be viewed as decision-theoretic generalizations of goal regression [4].

Reachability analysis allows one to determine that certain states are not reachable in an MDP given a particular initial state (or a set of possible initial states, or an initial distribution), no matter what actions are performed. This can be used to restrict dynamic programming to reachable states, reducing the computational burden of solving an MDP (for instance, see the envelope approach of [9]). However, approaches that determine reachability using explicit transition matrix operations [13] or state-based search cannot be exploited by variable-based abstraction techniques.

Our aim is to develop methods that determine the set of reachable states *implicitly* by explicitly considering the variable values or combinations of variable values that can or cannot be realized. This knowledge can be integrated into abstraction techniques by eliminating abstract states that contain unreachable variable combinations. The method we develop is based on the GRAPHPLAN algorithm [3]: the graph-building phase of GRAPHPLAN can be viewed as performing an approximate reachability analysis that is used to prune subsequent goal-regression search in a classical planning framework. However, we make certain modifications designed to deal with the DBN action representation. In particular, the distributed nature of this representation, while often more compact than (say) probabilistic STRIPS operators [15], requires that care be taken to deal with correlated effects. A second way in which we generalize the graph-building phase of GRAPHPLAN is to consider both simpler and more complex constraints in the construction of exclusion relations. We argue that reachability can be performed with varying levels of *completeness*, leading to one way of addressing anytime tradeoffs.

---
* This work was supported by NSERC Research Grant OGP0121843 and IRIS-II Project IC-7.

† Part of this work was undertaken at U.B.C. and was supported by NSERC.

‡ Part of this work was undertaken at U.B.C. and was supported by IRIS-II Project IC-7.

---
[1]The integration of reachability with state aggregation has been considered in the verification community (e.g., see [19]). There explicit state space representations are used however.

In Section 2, we review MDPs, DBN representations of MDPs, and briefly discuss techniques for policy construction that exploit the structure laid bare by this representation. In Section 3, we describe a family of algorithms based on GRAPHPLAN, where the complexity of exclusion constraints provides a certain latitude in the sophistication of the reachability analysis performed. We also consider the empirical performance of the algorithms and provide several results relating to soundness and completeness of the algorithms. In Section 4, we describe how the output of this analysis (in the form of variable value constraints) can be used to abstract MDPs, and describe some results illustrating the tradeoff between reachability sophistication and the size of the abstracted MDPs. We conclude in Section 5 with additional discussion.

## 2   MDPs and Their Representation

### 2.1   Markov Decision Processes

We assume that the system to be controlled can be described as a fully-observable, finite *Markov decision process* [1, 16], with a finite set of system states $S$. The controlling agent has available a finite set of actions $A$ which cause stochastic state transitions: we write $\Pr(s, a, t)$ to denote the probability action $a$ causes a transition to state $t$ when executed in state $s$. A real-valued reward function $R$ reflects the objectives of the agent, with $R(s)$ denoting the (immediate) utility of being in state $s$. A (stationary) *policy* $\pi : S \rightarrow A$ denotes a particular course of action to be adopted by an agent, with $\pi(s)$ being the action to be executed whenever the agent finds itself in state $s$. We assume an infinite horizon (i.e., the agent will act indefinitely) and that the agent accumulates the rewards associated with the states it enters.

In order to compare policies, we adopt *expected total discounted reward* as our optimality criterion; future rewards are discounted by rate $0 \leq \beta < 1$. The value of a policy $\pi$ can be shown to satisfy [16]:

$$V_\pi(s) = R(s) + \beta \sum_{t \in S} \Pr(s, \pi(s), t) \cdot V_\pi(t)$$

The value of $\pi$ at any initial state $s$ can be computed by solving this system of linear equations. A policy $\pi$ is *optimal* if $V_\pi(s) \geq V_{\pi'}(s)$ for all $s \in S$ and policies $\pi'$. The *optimal value function* $V^*$ is the same as the value function for any optimal policy. A number of state-based techniques for constructing optimal policies exist, including value iteration [1] and policy iteration [16].

### 2.2   Structured Representation and Computation

One of the key problems facing the use of MDPs for DTP is Bellman's "curse of dimensionality:" the number of states grows exponentially with the number of problem variables. Fortunately, several good representations for MDPs, suitable for DTP, have been proposed that alleviate the associated representational and computational burdens. These include stochastic STRIPS operators [12, 15] and dynamic Bayes nets [6, 10]. We will use the latter.

We assume that a set of variables $\mathbf{V} = \{V_1, \cdots V_N\}$ describes our system, with each $V_i$ having a finite domain $Dom(V_i)$ of possible values. To represent actions and their transition probabilities, for each action we have a *dynamic Bayes net* (DBN) with one set of nodes representing the system state prior to the action (one node for each variable), another set representing the world after the action has been performed, and directed arcs representing causal influences between these sets. Each post-action node has an associated *conditional probability table* (CPT) quantifying the influence of the action on the corresponding variable, given the value of its influences (see [6, 7] for a more detailed discussion of this representation). Figures 1(a) and (b) illustrate network fragments for two different actions (spray painting three parts, and assembling parts $P4$ and $P5$ into $P6$).

The lack of an arc from a pre-action variable $X$ (or a post-action variable $X'$) to a post-action variable $Y'$ in the network for action $a$ reflects the independence of $a$'s effect on $Y$ from the prior value of $X$ (or its effect on $X'$). We capture additional independence by assuming structured CPTs. In particular, we use a *decision tree* to represent the function that maps combinations of parent values to (conditional) probabilities. For instance, the tree in Figure 1(a) shows that $DryPx$ (is part $x$ dry) influences the probability of $PntPx$ (is $x$ painted) becoming true only if $PntPx$ is false and $MntPx$ (is $x$ mounted) is true (left arrows denote "true" and right arrows "false").[2] Arcs between post-action variables represent *correlated* action effects. For instance, in Figure 1(b), the effect of assembly on the ready status of $P4$ and $P5$ is determined by the success of assembly. Note that this representation is *distributed* in the sense that the effect of an action on distinct variables is captured by distinct CPTs. Instead of treating each possible effect as a conjunction of literals as in STRIPS, the DBN specifies the effects on each variable "independently." This offers economy of representation for many types of actions. The painting action, for instance, has three independent effects on three different parts. The representation of this action in (probabilistic variants of) STRIPS [15] requires specifying all eight combinations of painted/not painted values; and the size of the STRIPS representation generally grows exponentially with the number of independent effects.

A similar representation can be used to represent the reward function $R$, as shown in Figure 1(c). Here we see a fragment of a reward function that indicates that parts $P1$ and $P2$ *both* must be drilled to be worthwhile (reward 20)—with a slight cost incurred if drilling is done so as to cause wear—while an independent reward of 5 is received for assembling $P6$.

Apart from the naturalness and conciseness of representation offered by DBNs and decision trees, these representations lay bare a number of regularities and independencies that can be exploited in optimal and approximate policy construction. Methods for optimal policy construction can use compact representations of policies and value functions in order to prevent enumeration of the state space.

---

[2]Certain persistence relations can be exploited in the specification of actions: the dashed arcs (such as those for $MldPx$ and $MntPx$ in Figure 1(a)) correspond to persistence distributions where the variable retains its value after the action. We refer to [7] for a detailed discussion of persistence in DBNs.

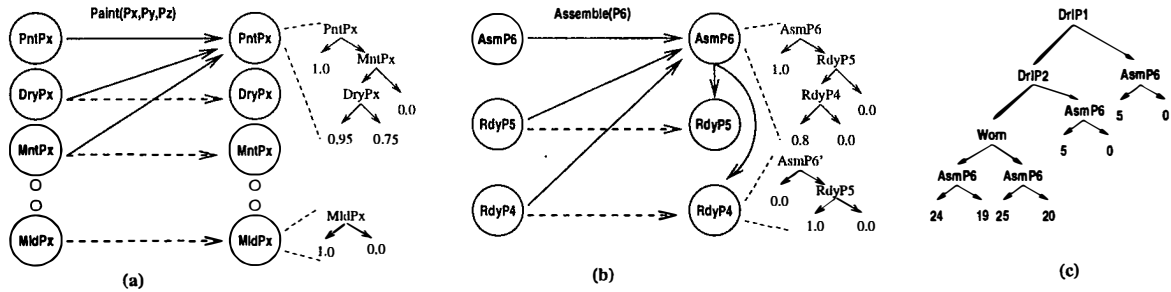

Figure 1: (a) Action network (spray painting); (b) action network with correlations (assembly); (c) reward tree

In [6] a structured version of modified policy iteration is developed, in which value functions and policies are represented using decision trees and the DBN representation of the MDP is exploited to build these compact policies. Roughly, the DBN representation can be used to dynamically detect the relevance of various variables under certain conditions at any point in the computation, thus allowing states to be aggregated by ignoring irrelevant variables. For instance, in the example described above, states where *AsmP6* holds are never distinguished by the truth values of *RdyP4* or *RdyP5*, since once *AsmP6* holds these facts are irrelevant to value or the optimal choice of action. The method can be extended to deal with approximation by using "degrees" of relevance as well [5].

A simpler abstraction technique developed in [12] uses a *static* analysis of the problem (as opposed to a dynamic analysis) to delete irrelevant or marginally relevant variables in the problem description. For instance, the fine differences in reward associated with *Worn* in Figure 1(c) can be ignored by deleting this variable. The DBN description of actions allows one to easily detect that certain other variables (for instance, careful alignment) have no impact, through any "causal chain" (or sequence of actions), on the truth of the remaining reward variables. An abstract MDP is formed by removing from the problem description any variables deemed *irrelevant* in this sense. The smaller MDP is easier to solve, but its solution will usually be suboptimal (though with easily determined error bounds).

## 3    Structured Reachability Analysis

One of the distinguishing features of MDPs *vis-à-vis* classical planning models is the need to construct policies that describe appropriate actions at all states. If we know the initial state (e.g., as in a planning problem), computational effort may be "wasted" in the determination of appropriate actions for states that can never be realized given the initial state, even if abstraction methods are used.

However, it is often the case that we can rather easily determine that certain variable values or combinations cannot be made true given the initial state. Any such knowledge of reachability can be exploited to reduce the range of dynamic programming. Furthermore, if this reachability analysis can be performed in a structured manner—that is, at the level of state variables rather than explicit states—the results can be combined with the policy construction methods and abstraction techniques described above. For instance, in our example above, if the initial state is known to lack sufficient material to produce either of *P4* or *P5*, then their ready status cannot be changed, nor can the state of assembly of *P6*. As a consequence, one can legitimately remove all mention of these variables from the MDP description—by fixing them to have the same values they have at the initial state—resulting in a *reduced MDP* of much smaller size (we discuss in detail how to construct a reduced MDP in the next section). If there is enough material to produce one of *P4* or *P5*, then *RdyP4* and *RdyP5* might change, but the condition on *P6* cannot, and again *AsmP6* can have its value fixed.

### 3.1    Reachability Analysis without Direct Correlations

Our algorithm for structured reachability analysis in inspired by GRAPHPLAN [3], a classical planning algorithm that essentially performs a reachability analysis to construct a *plan graph* and then performs goal regression within that graph. Specifically, the graph building phase of GRAPHPLAN operates by alternating the construction of *propositional levels* and *action levels*. At each propositional level is a set of propositions (or variable values) together with *exclusion constraints* among pairs of propositions that indicate that a state cannot be reached that makes both true. The initial state determines the first propositional level. An action level is created from the preceding propositional level by creating nodes for those actions whose preconditions are satisfied (reachable) at the previous level. Two actions are marked as exclusive if they "conflict" and the next propositional level is created by considering the effects of the actions at the preceding action level. Exclusions on these propositions are determined by examining conflicts between their producing actions.

Our algorithm follows the same general pattern as the graph building phase of GRAPHPLAN; in particular, we alternate the creation of action and propositional levels, and represent reachable states at the propositional level implicitly through the presence of variable values and exclusion constraints. Our algorithm is rather different in several respects, however, due to its application to MDPs and the nature of the DBN representation of actions. Key differences due to the DBN representation include the need to handle conditional effects, the distributed nature of action effects, and correlations that arise within actions or simply because of initial conditions. Because we are dealing with infinite horizon MDPs, we are interested only in the long-run reachable states, not in the transient behavior of reachable states. A final distinction is our emphasis on the tradeoff between arbitrary exclusion constraints and computational efficiency, in contrast to the binary constraints considered in GRAPH-PLAN. We elaborate on these differences below.

The algorithm REACHABLEK is sketched in Figure 2. The annotation $K$ refers to the fact that it deals with only $n$-ary constraints for $n \leq K$; this is, in fact, a family of algorithms, where the *complexity parameter* $K$ can range from 1 to $N$ (the number of domain variables). As we discuss below, the larger $K$ is, the more complete the resulting analysis, but the more complex the algorithm. The algorithm assumes that no actions have correlated effects (i.e., there are no "within slice" arcs). We do this for ease of presentation, describing the appropriate changes for correlated effects below.[3]

We begin with some preliminary definitions. Let $B$ be the DBN for an action $a$, let $V$ be some variable, and let $CPT(a, V)$ denote the tree quantifying $a$'s effect on $V$. $V$ is *unaffected* by $a$ (or *persists* under $a$), if $V$ retains its previous value under all conditions (these are captured by dashed persistence arcs). Otherwise $V$ is *affected* by $a$. For any affected variable $V$, each branch of $CPT(a, V)$ is a *condition* for $V$ w.r.t. $a$. The *effect set* of action $a$ on variable $V$ under condition $C$ is the set of variable values $\{v_i\}$ such that $\Pr(V = v_i | C, a) > 0$ (i.e., the set of values that $V$ *might* take if $a$ is executed when $C$ holds). An $n$-ary *exclusion constraint* over a set of variable values is any set of $n$ values drawn from that set.

The input to REACHABLEK is an initial state consisting of an assignment of values to all variables.[4] The output will be a set of variable values together with a set of $n$-ary *exclusion constraints*, $n \leq K$, over these values. We interpret this output as follows: all states consisting of some assignment of values drawn from the output set are (estimated to be) reachable unless this assignment contains some exclusion constraint. Thus a set of reachable states is represented (and constructed) in an implicit, structured fashion.

As with GRAPHPLAN, we alternate the construction of action and propositional levels. A propositional level is, as with the algorithm output, a set of variable values with $n$-ary exclusion constraints. The initial propositional level is determined by the initial state. The main loop shows how to construct an action level from the previous propositional level and how to construct the next propositional level from that action level.

**Action Level Creation:** An action level consists of *CAE nodes* and $n$-ary exclusion relations among them. A CAE node is a triple $\langle C, a, e \rangle$ where: $C$ is a *condition* or a conjunction of literals; $a$ is an action; and $e$ is an effect (variable value). Intuitively, this refers to the fact that action $a$ performed under condition $C$ might have effect $e$ (we say "might" because actions are stochastic). The first step of action level creation simply creates CAE nodes: for any action $a$ and variable $V$ affected by $a$, we create a CAE-node for each condition $C$ labeling a branch of $CPT(a, V)$ and effect (or value of $V$) $e$ in the effect set for $V$. However, this is only done for those conditions $C$ that are consistent at

---

[3] We will see that *induced correlations* can still arise and must be accounted for, even under this assumption.

[4] The extension to multiple initial states (e.g., an initial distribution) simply requires that the possible variables values and appropriate exclusion constraints be used as input.

---

Let *Values* = $\{v_j^i : i \leq N\}$ (initial state)
Let *Excl* = $\emptyset$
Loop until no change in *Values*, *Excl*
  1. Create Action Level:
    i. For each action $a$, affected variable $V$, condition $C$ labeling a branch of
      $CPT(a, V)$ such that $C$ occurs in *Values*, *Excl*, and value $v$ is in this effect set:
      Add $\langle C, a, v \rangle$ to action level.
    ii. For each $v$ in *Values*: add $\langle v, no\text{-}op(v), v \rangle$ to action level.
    iii. For each node $\langle C, a, v \rangle$ in action level, and each distinct variable $V'$
      affected by $a$, construct $IS(\langle C, a, v \rangle, V')$.
    iv. For each $\langle C_1, a_1, v_1 \rangle, \langle C_2, a_2, v_2 \rangle$ at the current action level: mark the
      pair as exclusive if $v_1$ conflicts with $C_2$ or $v_2$ conflicts with $C_1$, unless $a_1 = a_2$
    v. For any subset $S$ of action nodes, $n = |S| \leq k$, with conditions $C_1, \cdots C_n$:
      mark $S$ as exclusive if some subset of $\cup C_j$ is in *Excl*. $\{v_1, v_2\} \in Excl$; or
    vi. Repeat until no action exclusions found:
      If $\langle C_2, a_2, v_2 \rangle$ is marked as exclusive of each element of $IS(\langle C, a, v \rangle, V')$,
      then mark $\langle C_2, a_2, v_2 \rangle$ exclusive of $\langle C, a, v \rangle$
  2. Create New Proposition Level: *Values*, *Excl*
    i. For each $\langle C, a, v \rangle$ at current action level:
      Add $v$ to *Values*; record $\langle C, a, v \rangle$ as "a way to achieve" $v$;
    ii. For each set $S \subseteq$ *Values*, such that $|S| \leq K$:
      If all ways of achieving elements of $S$ have exclusive elements, add $S$ to *Excl*.

Figure 2: Algorithm REACHABLEK

the previous propositional level. In addition, as in GRAPH-PLAN, we assume that a special action $no\text{-}op(v)$ exists for every value $v$: its condition is simply $v$ and it has no effect other than $v$ (this deals with persistence relations). These are applied in the second step of action level creation.

One superficial difference with GRAPHPLAN is our use of conditions in action nodes: the fact that action $a$ has conditional effects is treated by creating a different "action" node (here a CAE node) for each condition under which $a$ has different effects.[5] GRAPHPLAN can easily be extended to deal with conditional effects in this way in classical settings (see [14, 18] where similar suggestions are made). A more substantial difference has to do with the distributed representation of action effects. To effectively deal with this, we create nodes for all possible independent effects of $a$, retaining the distributed flavor of the DBN model.

This tactic comes with its own difficulties however. Even under the assumption that actions do not have correlated effects, certain variable values may become correlated due to initial conditions. For example, imagine there is a light switch connected to 10 lights, $L0$ through $L9$. Suppose that toggling the switch deterministically flips any light from on to off, or off to on. Given any initial configuration of light values (on or off), there are only two reachable states. This cannot be captured using simple conflicts as in GRAPH-PLAN (see below).

To capture such *induced* correlations, we must examine the relation between the effects of a single action on a different variables. To do this, step iii. of action level creation determines the *implication sets* for each CAE-node. Let $\langle C_1, a, v_1 \rangle$ be a CAE-node, where $v_1$ is the value of some variable $V_1$ affected by $a$. For any distinct variable $V_2$ affected by $a$, we define $IS(\langle C_1, a, v_1 \rangle, V_2)$ to be the set of CAE-nodes $\langle C_2, a, v_2 \rangle$, where $v_2 \in Dom(V_2)$, such that $C_1$ and $C_2$ are consistent at the previous propositional level. Intuitively, if action $a$ is performed under condition $C_1$ and has effect $v_1$, it must also have *one* of the effects in $IS(\langle C_1, a, v_1 \rangle, V_2)$. In other words, $a$ cannot affect only one variable if the action in fact has several effects. Implication sets are confined to single actions, and can grow no larger than the size of the corresponding action description.

---

[5] If conditions for $a$ give rise to different distributions for the affected variable $V$ but have the same effect set (i.e., the same values have positive probability), these conditions need not be distinguished, since we are only concerned with whether values are *reachable*, not the probability with which they are reached. In Figure 1(a), the distinction *DryPx* is irrelevant to reachability. Preprocessing of DBNs to collapse such conditions is straightforward, and eases the complexity of reachability analysis.

Step iv. of action level generation proceeds almost exactly as GRAPHPLAN. We can execute a number of actions in "parallel" as long as they have consistent conditions and do not destroy each others effects or conditions. The one exception occurs when we have a single action with several distinct effects (which may destroy certain of the action's conditions). For example, if action $a$ has a condition (or precondition) $P$ that gives rise to two effects, $\neg P$ and $Q$, we do not want to mark the nodes $\langle P, a, \neg P \rangle$ and $\langle P, a, Q \rangle$ as exclusive simply because $a$ destroys one of its own conditions. We note that REACHABLE$K$ only ever considers *binary* exclusions of this form at the action level, due to the fact that the effects of actions at CAE-nodes are single literals, and the conditions of CAE-nodes are conjunctive. Thus, no set of actions can be in conflict (clobbering effects or preconditions) without some *pair* of actions within that set conflicting.

Step v. also tests for action exclusions based on the inconsistency of action conditions at the previous propositional level. Again, this is much like GRAPHPLAN; but in contrast to "clobbering" exclusions, here we must test $k$-ary action exclusions: the inconsistency of a set of action conditions cannot be reduced to pairwise tests. However, we can restrict attention to sets of actions of size $K$, since any detectable propositional exclusions can involve no more than $K$ values (thus exclusions of more than $K$ action conditions must be reducible to exclusions of a subset of size no greater than $K$).

The final stage of action level generation iteratively discovers additional conflicts through the use of implication sets. To illustrate, suppose $IS(\langle C_1, a, v_1 \rangle, V_2)$ describes the possible effects $a$ must have on $V_2$ when performed under condition $C_1$. If some other CAE-node $n$ (e.g., the effect of $a'$ on some other variable $V$) is marked as exclusive of *all* elements of this implication set, then $n$ must be exclusive of $\langle C_1, a, v_1 \rangle$. As an example, consider the light switch above, and suppose the initial configuration has $L0$ off and $L1$ on. The action nodes $\langle \mathit{off}_0, tgl, on_0 \rangle$ and $\langle on_1, no\text{-}op, on_1 \rangle$ do not directly conflict (by GRAPHPLAN's usual criterion); without accounting for implication sets we run the danger of judging both values to be simultaneously reachable (which is incorrect). However, $\langle \mathit{off}_0, tgl, on_0 \rangle$ has a singleton implication set for $L1$ consisting of $\langle on_1, tgl, \mathit{off}_1 \rangle$, which *does* conflict with $\langle on_1, no\text{-}op, on_1 \rangle$. Stage v. will detect this and mark the two CAE-nodes as exclusive. Thus, the induced correlation will be detected, and the fact that $L0$ and $L1$ have opposite parity will be maintained throughout the analysis.

**Propositional Level Creation:** The final part of the algorithm is propositional level creation. This is reasonably straightforward: we first create a value node for each variable value that is produced at the previous action level; we then mark any set of $n$ nodes as exclusive if any set of $n$ CAE-nodes that could produce the $n$ variable values contains a pair of CAE-nodes that is marked as exclusive at the previous action level. This requires a search through possible "assignments" of actions that produce these values to find a "satisfying assignment" (one that has no conflicting actions). While potentially all subsets of $n$ variable values must be investigated, we can generate subsets of increasing size to prune the search (any subset of values that includes a "smaller" exclusion constraint is itself infeasible). With no pruning, there can be as many as $2^k \sum_{i=1}^{N-k+1} i$ $k$-ary subsets of values (potential exclusion constraints) to be tested, if all variable values are present (assuming boolean variables). Thus, REACHABLE$K$ scales polynomially with all factors except the complexity parameter $k$.

**Termination:** The algorithm terminates when a fixed point is reached (i.e., when two consecutive propositional levels are identical). It is easy to verify that the algorithm will terminate: even if the underlying MDP exhibits periodic behavior, the presence of no-ops ensures that the set of states (implicitly) represented by each propositional level includes the set of states at the previous level. We also note that, unlike GRAPHPLAN, there is no need to keep track of any but the current action or propositional level. However, we could deal with finite horizon MDPs by making an explicit "planning graph" that stored all levels of the graph up to the horizon of interest.

## 3.2  Properties of the Algorithm

Two important properties of reachability algorithms are *completeness* and *soundness*. An algorithm is *sound* if every state considered unreachable by the algorithm is, in fact, unreachable (or equivalently, all reachable states said to be reachable by the algorithm). This is important for accurate solution of an MDP: if all reachable states are included in the reduced MDP, the optimal policy for the reduced MDP will be an accurate reflection of optimal behavior with respect to the given initial state. More specifically, let $\widehat{M}$ denote a reduced version of MDP $M$ obtained by removing all states deemed unreachable by a reachability algorithm $A$. Let $\widehat{S} \subseteq S$ be the state space for $\widehat{M}$, let $\widehat{\pi}*$ be an optimal policy for $\widehat{M}$, and let $V^*$ denote the optimal value function for $\widehat{M}$. If $A$ is sound, we are assured that:

**Proposition 1** *Let $\pi$ be any policy for $M$ that extends $\widehat{\pi}*$ (i.e., $\pi(s) = \widehat{\pi} * (s)$ for any $s \in \widehat{S}$). Then, for any $s \in \widehat{S}$, we have:* $V_{\widehat{\pi}*}(s) = V_{\pi}(s) = V^*(s)$

An algorithm is *complete* if all unreachable states are said to be unreachable by the algorithm, i.e., all unreachable states are recognized. Completeness ensures that no unreachable states are included in the reduced MDP, and it has the effect of keeping the reduced MDP small, though, on its own, does not guarantee an optimal solution.[6]

Recall that the output of REACHABLE$K$ is the set of variable values at the last propositional level together with $n$-ary exclusion constraints over those values, $n \leq K$. For any $1 \leq K \leq N$, the output of REACHABLE$K$ is sound:

**Theorem 2** *If state $t = \langle v_{i_1}^1, v_{i_2}^2 \cdots v_{i_N}^N \rangle$ is reachable from initial state $s$, then each value $v_{i_k}^k$ that makes up state $t$ is in*

---

[6]Notice that the terms *sound* and *complete* are w.r.t. statements of *un*reachability, which is really what we are interested in.

*the set of values returned by* REACHABLEK. *Furthermore, no subset of values making up* t *is marked as exclusive.*

Finally, it is easy to see that the more stringent the exclusion constraints, the fewer states will be deemed reachable.

**Proposition 3** *If* $S_K$ *is the set of reachable states returned by* REACHABLEK *and* $S_J$ *is the set of reachable states returned by* REACHABLEJ *(for a fixed initial state* s*), where* $J > K$ *then* $S_J \subseteq S_K$.

In other words, stronger constraints lead to more complete reachability analysis.

Unless REACHABLEK produces identical results for all values $K \leq N$, it is clear that REACHABLEK cannot generally be complete. In fact, it is not hard to see why this is the case. Consider the example action in Figure 1(a), where three parts are spray painted and imagine that it can be instantiated with any three parts from the set {*P1, P2, P3, P4*}. Furthermore, imagine that three litres of paint are consumed by the action. If the initial state is such that there are exactly three litres of paint, we have three possible instantiations of the paint action that can be executed (ignoring permutations): *Paint(P1, P2, P3)*, *Paint(P1, P2, P4)*, and *Paint(P2, P3, P4)*. Clearly each pair of these will be marked as exclusive at the first action level. If we are implementing REACHABLE2, we consider only *pairs* of values when checking exclusions at the subsequent proposition level; but no pair of propositions chosen from {*PntP1, PntP2, PntP3, PntP4*} will be marked as exclusive—there exists a single action that can produce any pair. This means that no exclusion constraints will be generated at the propositional level. When the algorithm terminates (assuming nothing else influences these propositions), we are left to assume that a state in which all four parts are painted can be realized, which is incorrect. Thus while soundness is guaranteed (hence, "correct" solution of the reduced MDP is assured), completeness does not hold, meaning the reduced MDP may be larger than necessary.[7]

We note that GRAPHPLAN deals only with binary constraints (and can be viewed as a special form of REACHABLE2) and is thus incomplete in its analysis. Again, this incompleteness does not impact the correctness of planning, just (potentially) the complexity of planning. In general, completeness is not assured except for REACHABLEN, where $N$ is the number of domain variables.

**Theorem 4** *If state* $t = \langle v_{i_1}^1, v_{i_2}^2 \cdots v_{i_n}^n \rangle$ *is consistent with the output of* REACHABLEN *given initial state* s*, then* t *is reachable from* s.

This completeness guarantee comes at high computational cost. Testing every set of $N$ values at a propositional level induces a huge combinatorial explosion: at worst, if every value of every variable is realizable (even if not in combination), one ends up enumerating the entire state space (recall that the algorithm grows exponentially with the complexity parameter $k$). However, domain or operator specific knowledge may be used to restrict the size of $k$ in certain circumstances. Furthermore the monotonicity guarantee of Proposition 3 suggests that the family of reachability algorithms can be used in an anytime fashion to do more detailed reachability analysis as time permits.

---

[7]Note that in this example quaternar constraints are needed to prove that all four parts cannot be painted.

| | Alg. | Rch. MDP | Eff. MDP | Time (s) |
|---|---|---|---|---|
| | Rch1 | $2^{16}$ | 0 | 0.02 |
| s1 | Rch2 | $2^{16}$ | 0 | 1.19 |
| | Rch3 | $2^{16}$ | 0 | 62.96 |
| | Rch1 | $2^{17}$ | $2^2$ | 0.02 |
| s2 | Rch2 | $2^{17}$ | $2^2$ | 1.21 |
| | Rch3 | $2^{17}$ | $2^2$ | 68.02 |
| | Rch1 | $2^{20}$ | $2^4$ | 0.02 |
| s3 | Rch2 | $2^{18}$ | 0 | 1.68 |
| | Rch3 | $3/4 \cdot 2^{18}$ | 0 | 109.25 |
| | Rch1 | $2^{21}$ | $2^{19}$ | 0.04 |
| s4 | Rch2 | $97/128 \cdot 2^{19}$ | $97/128 \cdot 2^{19}$ | 183.76 |
| | Rch3 | $1067/2048 \cdot 2^{19}$ | $5/8 \cdot 2^7$ | 6744.43 |
| | Rch1 | $2^{25}$ | $2^{25}$ | 0.06 |
| s5 | Rch2 | $97/128 \cdot 2^{21}$ | $97/128 \cdot 2^{17}$ | 194.12 |
| | Rch3 | $3201/4096 \cdot 2^{20}$ | $5/8 \cdot 2^7$ | 8019.12 |
| | Rch1 | $2^{26}$ | $2^{26}$ | 0.04 |
| s6 | Rch2 | $97/128 \cdot 2^{22}$ | $97/128 \cdot 2^{18}$ | 197.59 |
| | Rch3 | $3201/4096 \cdot 2^{21}$ | $5/8 \cdot 2^8$ | 8172.19 |

Table 1: Results of REACHABLEK ($K = 1, 2, 3$) for different initial states

**Empirical Evaluation:** To illustrate typical behavior, we have run REACHABLE1, REACHABLE2 and REACHABLE3 on a small manufacturing domain. Space precludes a full domain description, but roughly, we have six parts that play a role in the reward function: value is attached to performing certain operations on these parts. Some of the objectives are conjunctive, for example, processing parts *P1* and *P2* (in this case, painting, drilling and polishing) only has value if *P3* is also processed.[8] The domain is described by 31 binary propositions (with $2^{31}$ states, far larger than can be handled by explicit MDP solution algorithms) and 30 actions (including a spray paint action for two parts, similar to that in Figure 1(a)).

We have run each algorithm on six different initial states that intuitively vary in the number of states they render reachable. This is achieved by varying the action preconditions and resource constraints at the initial state. For instance, $s1$ (in Table 1) meets no "useful" action preconditions and has no resources available for building or assembling parts. Other states allow certain objectives to be met, but not others; for instance, in $s4$, $s5$ and $s6$ there is not enough paint to paint more than two parts.

The results are summarized in Table 1, where for each state-algorithm pair we list the size of the *reachable MDP* given that initial state as estimated by the algorithm, the size of the *effective MDP* (described in the next section), and the time taken to construct the reachable set.[9]

---

[8]This relation between value-laden propositions will be important in the next section.

[9]The algorithm was implemented in Prolog and run using Sicstus Prolog on a Sparc Ultra2. The algorithm implemented has few optimizations, except that $k$-ar constraints are generated in increasing order of $k$, allowing some "pruning." Rather than test for termination, here we simply ran REACHABLEK to three levels, since in all examples a fixed point is reached at or before the third propositional level.

We see that, as expected, REACHABLE3 determines a smaller (estimated) reachable region than REACHABLE2, which in turn deems fewer states reachable than REACHABLE1. For instance, at state $s1$, where no "useful" resources or preconditions are met, these algorithms all quickly discover that 15 of the 31 variables cannot change value, resulting in a reachable MDP with size $2^{16}$. Here no binary or ternary exclusion constraints exist, thus all three algorithms produce the same results. In the final four states, the algorithms differ, with more complicated processing resulting in a smaller reachable MDP being determined. Notice that the amount of time required by REACHABLE2 and REACHABLE3 increases dramatically with the "complexity" of the reachable MDP (i.e., the number of variable values realizable and actions executable), with REACHABLE3 taking on the order of 2 hours for the last two problems. While the sizes of the "reachable MDPs" may still prove too unwieldy for effective solution, these results can be used to leverage abstraction techniques, as we describe in the next section, making the effective MDP more manageable still.

Apart from considering $k$-ary constraints for smaller values of $k$, there are a number of other shortcuts one can adopt in the algorithm that ensure soundness while sacrificing a certain degree of completeness. Among these are: performing an incomplete search for $k$-ary exclusion constraints (which might allow unreachable value combinations to be designated reachable); and performing partial propagation of action exclusions through implication sets. These make the algorithm quite flexible for the purposes of anytime computation; and none of these adjustments affect the ability to accurately solve the reduced MDP (but do affect its size and, hence, required solution time). A more incremental variant of this model with potential anytime applicability could also be adopted in which REACHABLEK is run on successively larger values $K$. The algorithm can be modified so that $n$-ary constraints are computed using the output of the $n-1$-ary output.

### 3.3 Including Correlated Action Effects

The algorithm REACHABLEK presented above is designed to deal with DBNs representing actions without correlated effects. The action depicted in Figure 1(b), to give one example, would not be dealt with correctly. We note, however, that simple probabilistic dependency among action effects does not cause difficulties: it is only *deterministically correlated* effects that are problematic. More precisely, if the effect of an action on one variable depends on its effect on another, such that the set of *possible values* (i.e., values with positive probability) taken by the second variable vary with the action's effect on the first, we call these deterministically correlated. This restricts the value combinations that the action can produce. Correlations that reflect simple changes in the probabilities of certain values, but not the possible values that can be realized, will be ignored. In the remainder of this section, we use the terms correlation and dependence in this stronger sense.

One method of coping with correlated effects is to construct *compound variables*: any variables in a DBN for which action $a$'s effects are dependent are joined, forming a single variable taking as values assignments to the individual variables. Only the effect on post-action variables are merged: the pre-action variables can stay as they were. The distribution over effects on this compound variable is then easily computed using the original DBN. Different DBNs will typically have different groupings of variables (if any).

REACHABLEK can work directly on this modified representation, requiring only minor adjustments. When CAE nodes are defined, they may refer to compound variables. Testing for action conflicts requires comparing not just individual variables, but also compound variables. However, this is straightforward, involving comparing lists of literals for conflicts. Creating the propositional level requires that we "split" any compound effects of CAE nodes. But no special attention is required to create exclusion constraints at the propositional level, since constructing implication sets and action conflicts using compound variables will impose any dependencies on the propositional level. Our current implementation of REACHABLEK works in this fashion.

One disadvantage of creating compound variables is the potential for exponential blowup in the domain sizes of such variables. We are currently exploring a version of REACHABLEK that deals directly with correlations in action and proposition level construction. We defer details to a longer version of this paper; but roughly we can use the structure of the network, specifically the dependencies among action effects, to carefully construct implication sets for different variables. Once the implication sets are correctly constructed, the algorithm works much like it does currently. The main distinction lies in the fact that the conditions in CAE nodes can now refer to post-action variables in certain cases. This requires some care in detecting conflicts among actions. We hope to examine empirically the differences in performance between this strategy and our current (compound variable) approach in the near future.

We note that one can simply ignore correlated effects in the algorithm. As with the shortcuts mentioned in the previous section, this will not affect the soundness of the algorithm, but can cause the set of reachable states to be overestimated.

## 4    Abstraction and Reachability Combined

The most important aspect of our reachability algorithms is that they produce a representation of the (estimated) reachable states in a structured form—a set of variable values with exclusion constraints relating them. We refer to this output as the *reachable set*. Because REACHABLEK does not produce an explicit list of reachable states, it can be integrated readily with MDP representations such as DBNs and the abstraction algorithms for MDPs that were briefly discussed in Section 2.2. In fact, there are several ways in which this representation of reachable states can be exploited by these abstraction techniques. We describe a sequence of increasingly sophisticated ways in which to exploit reachability constraints.

The simplest way in which to use the reachable set produced by REACHABLEK is to remove any variable values from the MDP description that do not occur in the reachable set. For instance, if $V$ has only two possible values in the reachable set, say, $v_1$ and $v_2$, all other values can be ignored, since they cannot be realized. Furthermore, a variable $V$ is completely *removable* if it has only one reachable value. To exploit this fact in the algorithms that use DBN representations of MDPs, we want to *reduce* the DBNs (and reward tree) of the MDP. This process is reasonably straightforward. Intuitively, we remove any unreachable variables values from the reward tree or CPT-trees by removing any edges in the tree labeled with unreachable values. If this results in a node with only one outgoing edge, the variable itself is removable (i.e., has only one reachable value), and is deleted from the tree (the subtree attached to the remaining edge is promoted). Any removable variables can be completely deleted from the DBNs as well. The reduction of this MDP representation can be performed in linear time, and results in reward tree and set of DBNs that accurately reflects the reduced or *effective* MDP. By retaining the structured nature of the representation, the result can be used by any standard abstraction algorithm (Section 2.2).

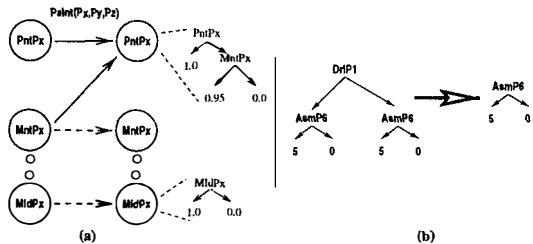

Figure 3: Reduced (a) action network and (b) reward tree

As an illustration, referring to Figure 1(a), we may discover through REACHABLEK that *DryPx* is always true. The DBN in that figure can then be reduced by removing any mention of the variable *DryPx* and replacing the CPT entry that refers to *DryPx* with the distribution determined by its "true" value—this is illustrated in Figure 3(a). Since *DryPx* now has a fixed value, it can, in fact, be completely removed from the problem description.

More complex processing of the same type can use the exclusion constraints in the reachable set. Rather than reducing trees and DBNs using only the absence of certain values, the fact that *combinations* of variable values cannot be realized can be exploited. In particular, we can prune any (reward or DBN) tree in which a branch is labeled with values that include an exclusion constraint. To do this requires that we simply prune that branch at the earliest point a constraint is violated. For example, referring to the reward tree in Figure 1(a), we may discover a binary exclusion constraint involving *DrlP1* and *DrlP2* (e.g., there is not enough skilled labor to make both true). The reward tree can then be reduced rather dramatically as shown in Figure 3(b): the infeasibility of certain objectives lets us dismiss certain others.[10]

Because our reachability techniques can be used to produce compact DBN representations of the reduced MDP, one can directly apply structured abstraction techniques to the reduced MDP in order to solve it. The advantage of first producing a reduced MDP is that the descriptions are generally smaller (and can certainly be no larger). This generally results in fewer distinctions being made by the structured algorithm being used. If one needs only solve a planning problem (for a given initial state or set of initial states) rather than the complete policy construction problem, this can offer a considerable advantage.

The simple abstraction algorithm of [12] can benefit substantially from this sort of analysis. As mentioned above, we may easily determine that certain objectives are infeasible and remove them from the problem description (e.g., *DrlP1, DrlP2* above). Removal of such variables can induce the abstraction algorithm to remove certain other variables as well: propositions that may have been relevant because they made the achievement of an objective more or less likely may be deemed irrelevant in the reduced MDP if the objective has been deleted. Thus, reachability analysis can lead to dramatically smaller abstract MDPs: not only are the unreachable values removed, but factors relevant only to such removed values can also be deleted.

**Empirical Evaluation:** In our manufacturing example described in the previous section, we get very substantial reductions in MDP size by first removing unreachable values and value combinations from the MDP description. Using the algorithm of [12] without reachability, assuming that only strictly irrelevant variables are removed (i.e., with no approximation), we can remove 2 of the 31 variables, reducing the MDP to size $2^{29}$. Integrating the output of REACHABLEK with abstraction provides much more substantial reductions in MDP size, as illustrated in Table 1, where *effective MDP* size shows how this size varies for different degrees of reachability and different initial states. All of these states lack certain useful resources or make certain key preconditions false; thus the *reachable* set of objectives at each state is restricted in one way or another.

For example, when REACHABLE1 is applied to state $s1$, although we find that many states are still reachable, we discover that none of our objectives (variables influencing reward) are within our control. The effective MDP has size zero, indicating that there is really no decision problem to be solved. Thus a simple reachability analysis prevents one from solving an MDP of size $2^{31}$. Other initial states (carefully chosen to make particular subsets of our objectives achievable) determine reduced MDPs with substantial reductions in size as well. Notice that at state $s2$, REACHABLE1 produces an effective MDP with only four states. At $s3$, REACHABLE1 cannot detect the fact that one simple objective is unachievable (resulting in a small MDP for that objective only), but REACHABLE2 discovers that even it is not reachable. States $s4$ through $s6$ involve an objective whose infeasibility can only be discovered using ternary exclusion constraints. Though a high overhead is involved in invoking REACHABLE3, the reduction in the sizes of the effective MDPs are dramatic when compared to the results of REACHABLE1 or REACHABLE2. At state $s4$, for instance, the effective state space is reduced by a factor of roughly 5000 when REACHABLE3 is used instead of REACHABLE2, potentially making a computationally infeasible decision problem tractable.

These results are illustrative and encouraging (though there is no reason in general to expect greater *marginal* reductions as the complexity of exclusion constraints increases—this is an artifact of our example). The fact that even simple reachability analysis can substantially reduce the effective size of an MDP seems clear; and more complex analysis will, at higher computational cost, provide deeper overall reductions.

---

[10] Here we prune the tree at node *DrlP2*, replacing it with the false subtree. Since the result is a tree with *DrlP1* at the root, but identical left and right subtrees, we take the liberty of removing the root node: clearly *DrlP1* is no longer relevant to reward. Generally such isomorphisms can be detected readily given canonical variable orderings.

The more sophisticated techniques of [4, 5, 6] can, of course, benefit in the same way as the simple abstraction method. However, these algorithms create dynamic, nonuniform abstractions, creating and recreating tree representations of value functions and policies. As such, the simple approach of using the reachable set to reduce MDP descriptions once and for all does not take full advantage of the reachability analysis. Algorithms like these can be augmented slightly to *fully* exploit the output of REACHABLEK: since these algorithms put together combinations of variables dynamically, they may start to compute policy values for states satisfying variable values that are marked as exclusive. It is a simple matter to add tests that use the constraints returned by REACHABLEK to rule out unrealizable variable value combinations when building policy and value trees. Roughly, when a tree is constructed that contains a branch labeled with variables values that are exclusive, a suitable subtree can be removed from the tree. This results in smaller trees being maintained, which is the main factor in reducing the complexity of these algorithms.[11] Once again, the fact that reachability analysis is performed using the structured representation of the MDP is the key to integrating reachability with abstraction.

## 5    Concluding Remarks

We have described an algorithm for performing structured reachability analysis of an MDP, using DBN action representations, that allows one to trade off the complexity of the analysis with completeness. Our approach is inspired by the GRAPHPLAN algorithm, but generalizes the concepts introduced there by handling conditional effects and correlations, and by providing the flexibility to adjust the degree of reachability considered. We have illustrated how this can be exploited in certain MDP abstraction algorithms to provide deeper MDP reductions when initial states are fixed. One cannot expect reachability to play a substantial role in MDP reduction in all cases (ergodic MDPs, for example); but it can be significant in many circumstances, such as when rewards are conditional, when certain variables are observable, but uncontrollable (e.g., road conditions, the weather, interest rates), or when resource constraints restrict the set of reachable states. Our preliminary empirical results on resource-constrained problems suggest that reachability can offer dramatic MDP reductions in certain circumstances.

A number of interesting directions emerge for future research. The application of these ideas to classical planning settings, when viewed as extensions to GRAPHPLAN, could prove useful. We are also interested in exploiting the distributional information in our action descriptions to perform approximate reachability analysis, focusing attention on the most "important" reachable states (e.g., perhaps the most probable, but accounting for improbable states that have a high impact on value).

**Acknowledgements:** Thanks to Bob Givan and Matthew Spears for their comments and for pointing out a correction to an earlier version of the algorithm.

## References

[1] Richard E. Bellman. *Dynamic Programming*. Princeton Uni-

---

[11]We defer full details to a longer version of this paper.